\documentclass[11pt,letterpaper]{article}
\usepackage{naaclhlt2016}
\usepackage{times}
\usepackage{latexsym}

\usepackage{url}
\usepackage{multirow}

\usepackage{latexsym}
\usepackage{graphicx}
\usepackage{amssymb}
\usepackage{subfigure}
\usepackage{multirow}
\usepackage{graphicx}
\usepackage{epstopdf}
\usepackage{amsmath}
\usepackage{CJKutf8}

\naaclfinalcopy 
\def\naaclpaperid{***} 

\title{A Planning based Framework for Essay Generation}

\author{Bing Qin, Duyu Tang\thanks{\ \ \  indicates equal contribution.} , Xinwei Geng$^*$, Dandan Ning$^*$, Jiahao Liu$^*$, Ting Liu \\
	    Harbin Institute of Technology, Harbin, China\\
	    {\tt \{qinb, dytang, xwgeng, ddning, jhliu,tliu\}@ir.hit.edu.cn}  }

\date{}

\begin{document}

\maketitle

\begin{abstract}
Generating an article automatically with computer program is a challenging task in artificial intelligence and natural language processing. 
In this paper, we target at essay generation, which takes as input a topic word in mind and generates an organized article under the theme of the topic.
We follow the idea of text planning \cite{Reiter1997} and develop an essay generation framework. 
The framework consists of three components, including topic understanding, sentence extraction and sentence reordering. 
For each component, we studied several statistical algorithms and empirically compared between them in terms of qualitative or quantitative analysis. 
Although we run experiments on Chinese corpus, the method is language independent and can be easily adapted to other language. 
We lay out the remaining challenges and suggest avenues for future research. 
\end{abstract}

\section{Introduction}
In general, natural language processing tasks could be divided into natural language understanding and natural language generation \cite{Manning1999,Jurafsky2000}. The former takes as input a piece of text and outputs the syntactic/semantic/sentimental information involved in the text. The latter in contrast focuses on generating a piece of text from an idea in mind or from a large collection of text corpus.
In this paper, we focus on natural language generation \cite{Reiter2000}.
Specifically, we formulate the task as essay generation from mind, namely taking the input as a topic word\footnote{Supposing the input topic word is unambiguous.} in mind and outputing an organized article (a document) with several paragraphs under the theme of the topic.
The task is challenging as it requires the generator to deeply understand the way human beings write articles.
Hopefully, solving this problem contributes to making progress towards Artificial Intelligence.

We argue that generating a well organized article is a challenging task. 
The first challenge is how to understand and represent the meaning of a topic word in mind. 
This is extremely important as telling the computer what we want to write is the first step we need to do before generating an article. 
Computer program does not have background like human beings, so that it does not understand a ``cellphone'' is an electronic product including battery and it can be used to chat with others. 
After understanding the meaning of a topic word, the following challenge is how to generate a topic focused article, e.g. how to collect topic-specific ``fuel'' (e.g. sentences) and how to organize them to form an organized article. 
This is of great importance as an article is not a set of sentences chaotically. Natural language is structured \cite{Mann1998,Jurafsky2000}. 
The coherence/discourse relationship \cite{Prasad2008,Li2014a,Li2014b} between sentences is a crucial element to improve the readability of a document and to guarantee the structured nature of a document in terms of lexicalization and semantic. 


\begin{figure*}[t]
	\centering
	\includegraphics[width=.7\textwidth]{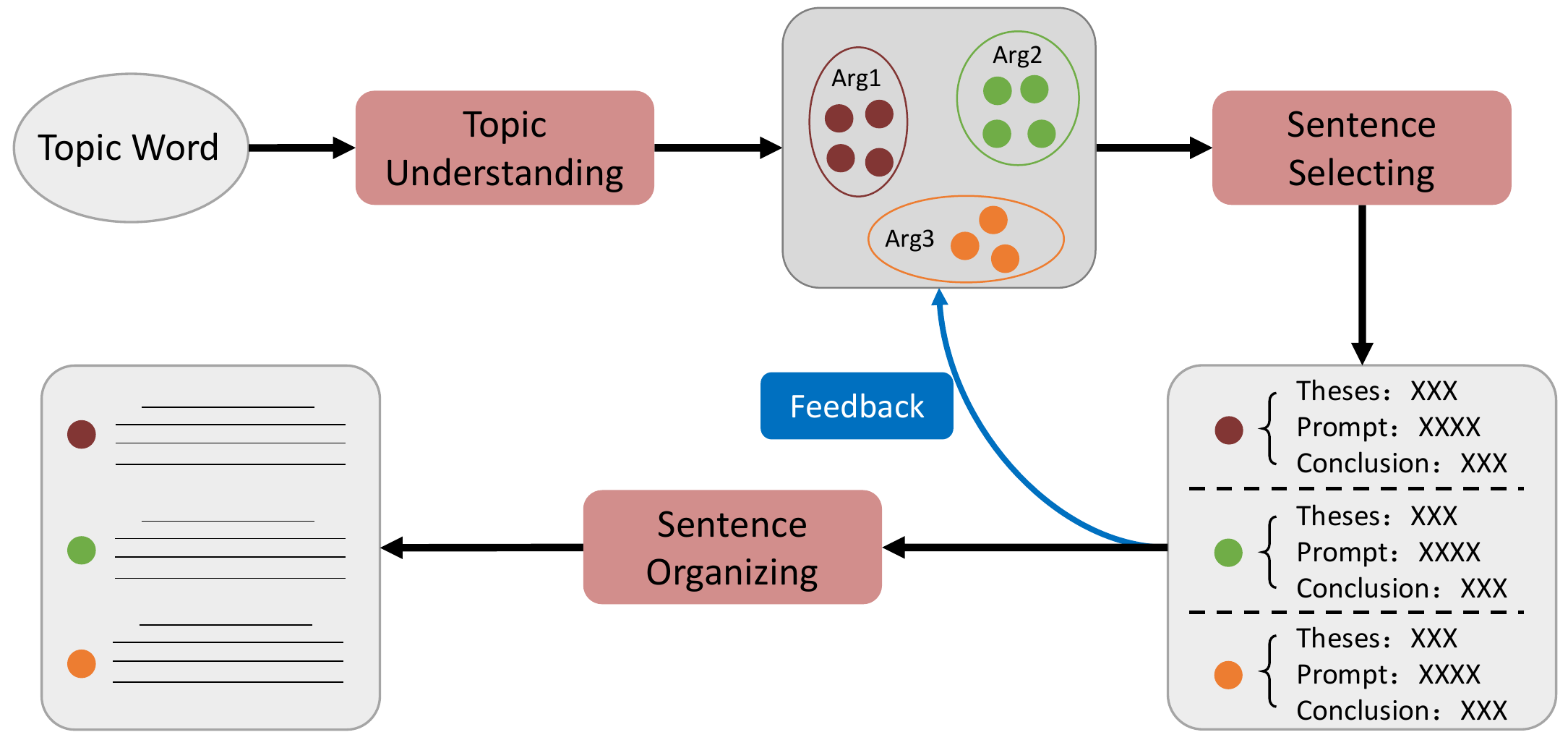}
	\caption{The planning based framework for essay generation.}
	\label{fig:framework}
\end{figure*}


In this paper, we develop a planning based framework \cite{Reiter1997,Reiter2000} for essay generation. 
The framework consists of three steps: topic understanding, sentence selecting and sentence organizing. 
Firstly, it represents the topic word in semantic vector space and automatically recognizes several arguments in order to support the topic. 
Each argument is represented as a list of supporting words which are semantically related to the topic word from a certain perspective.
Afterwards, a set of semantically related sentences are extracted/ranked for each argument given the list of supporting words.
Finally, the chaotic sentences with regard to each argument are organized to output an article by taking into account of the discourse and semantic relatedness between sentences. 
Furthermore, in order to find new evidences (e.g. words) to better support an argument, we add a feedback component to find new words from the extracted sentences to expand the existing evidence set. 

We conduct a case study on a Chinese corpus. 
For each component in the framework, we explore several strategies and empirically compare between them in terms of qualitative or quantitative analysis. 
We analyse the pros and cons of each approach, lay out the remaining challenges and suggest avenues for future research. 


\section{The Framework}
We describe the planning based framework in this section. 
As illustrated in Figure \ref{fig:framework}, the framework consists of three steps: topic understanding, sentence selecting and sentence organizing. 
We also add a feedback mechanism to enrich the supporting words of each argument.
We describe the details of these components, respectively.

\subsection{Topic Understanding}
When a person writes an article under the theme of a certain topic, he/she typically finds some arguments to support his main idea. For example, an article about ``cellphone'' might have three paragraphs, stating the evaluations towards ``call quality'', ``appearance'' and ``battery life'', respectively. 
These arguments are some important characteristics of the topic from some aspects.
The evidences about each argument make the whole article cohesive. 
Based on these considerations, we regard topic understanding as the first component of the framework. 
Given a topic word as input, topic understanding analyzes its semantic meaning and outputs several arguments to support the topic. Each argument is represented as a collection of words, each of which is semantically related with the topic from some aspect. 

\begin{figure}[h]
	\centering
	\includegraphics[width=.49\textwidth]{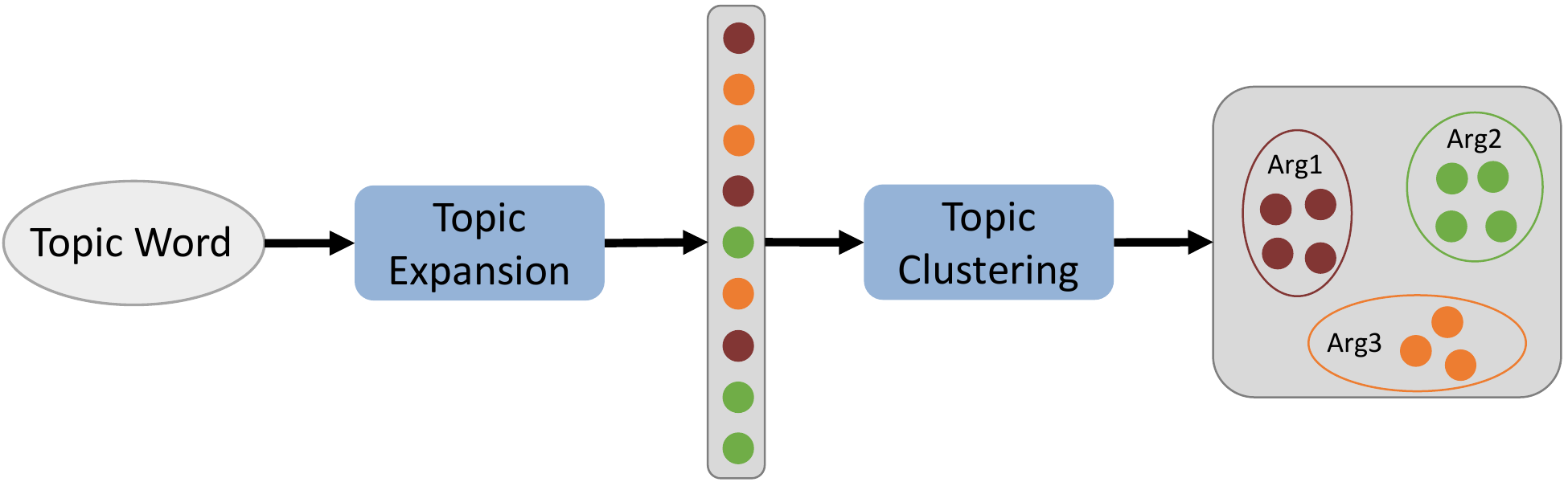}
	\caption{The pipeline for topic understanding.}
	\label{fig:topic-understanding}
\end{figure}

We consider topic understanding as two cascaded steps: topic expansion and topic clustering. An illustration is given in Figure \ref{fig:topic-understanding}. The former step finds a collection of words having similar semantic meanings with the topic word. The latter step separate similar words into several clusters, each of which share some properties with the topic word in terms of some aspects. 

Specifically, for the topic expansion component, we exploit thesaurus based, topic model based and word embedding based approaches.
Using external thesaurus is a natural choice as thesaurus like WordNet (or HowNet in Chinese) mostly contain the synonyms, antonyms, hypernym relationships. Accordingly, we can use heuristic rules to find more semantically related words as candidates. 
Straight forward rules include the synonym/hypernym of a word  or the antonym of an antonym.
Since some results might have noises, we can design a scoring function (e.g. the number of times a word occurs) to filter out some words with lower confidence. 
We also try a propagation strategy, where the extracted word set is further regarded as seeds and used to find more related words. 
Topic modeling and word embedding approaches  represent a word as a continuous vector in semantic vector space. 
Let us take word embedding as an example. Words with similar semantic meanings and grammatical usages will be mapped into closed vectors in the embedding space \cite{Mikolov2013a}. Therefore, to find some semantically similar words with the topic word, we could collect the neighboring words of the topic word in the vector space in terms of some similarity criterion like cosine or Euclidean distance with the topic word. We set a threshold to remain the most confident $k$ words and filter out the others.

After find a set of related words, we use standard clustering algorithms like K-Means and Affinity Propagation \cite{Frey2007} to separate words to several clusters, each of which represents an argument to support the topic. 
For topic model and word embedding approaches, the inputs of a clustering algorithm are the continuous representation of words. 
We pre-define the number of clusters of K-Means, while AP approach could automatically decide the number of clusters.

\subsection{Sentence Selecting}
After obtaining several clusters of words, each of which supports the topic word from an aspect/argument, we select a number of sentences for every argument. We can reuse a sentence selecting module several times to find evidences for each argument, respectively. 
Formally, for each argument, sentence selecting takes as input a collection of words and outputs a list of sentences with regard to the semantics of these words. The selected sentences will be used to compose a paragraph with sentence organizing, which is described in the following subsection.

Sentence selecting could be regarded as a retrieval problem, namely selecting the sentences with high similarities with a collection of words. 
Accordingly, defining a good scoring function $f(W, s)$ plays an important role to obtain a good result, where $W$ is a collection of words and $s$ is a sentence to be scored.   
We explore two kinds of methods in this work, a counting based method and an embedding based method. 
Counting based approach is a straight forward way to score each sentence, which is similar with the word matching strategy in information retrieval \cite{Manning2008}. 
The assumption is that a sentence $s$ containing more words in $W$ should be semantically closer to $W$. 
The scoring function $f(W, s)$ is the number of words in $W$ occurs in the sentence $s$, the higher the better. 
However, it is commonly accepted that a word typically has different semantic meanings and one meaning could be expressed by different word surfaces. Therefore, it is more desirable to develop a semantic driven approach, where $W$ and $s$ are mapped in the semantic vector space. 
Towards this goal, we consider an embedding-based approach, which maps $W$ and $s$ in a latent semantic space.
Accordingly,  we can compute the similarity between $W$ and $s$ in the embedding space. 
We use a simple average method as a case study in this work. 
We first map each word in a low-dimensional word embedding \cite{Mikolov2013a}. Afterwards, the representation of $W$ is the average of the vectors of words $w \in W$. 

\begin{figure}[h]
	\centering
	\includegraphics[width=.41\textwidth]{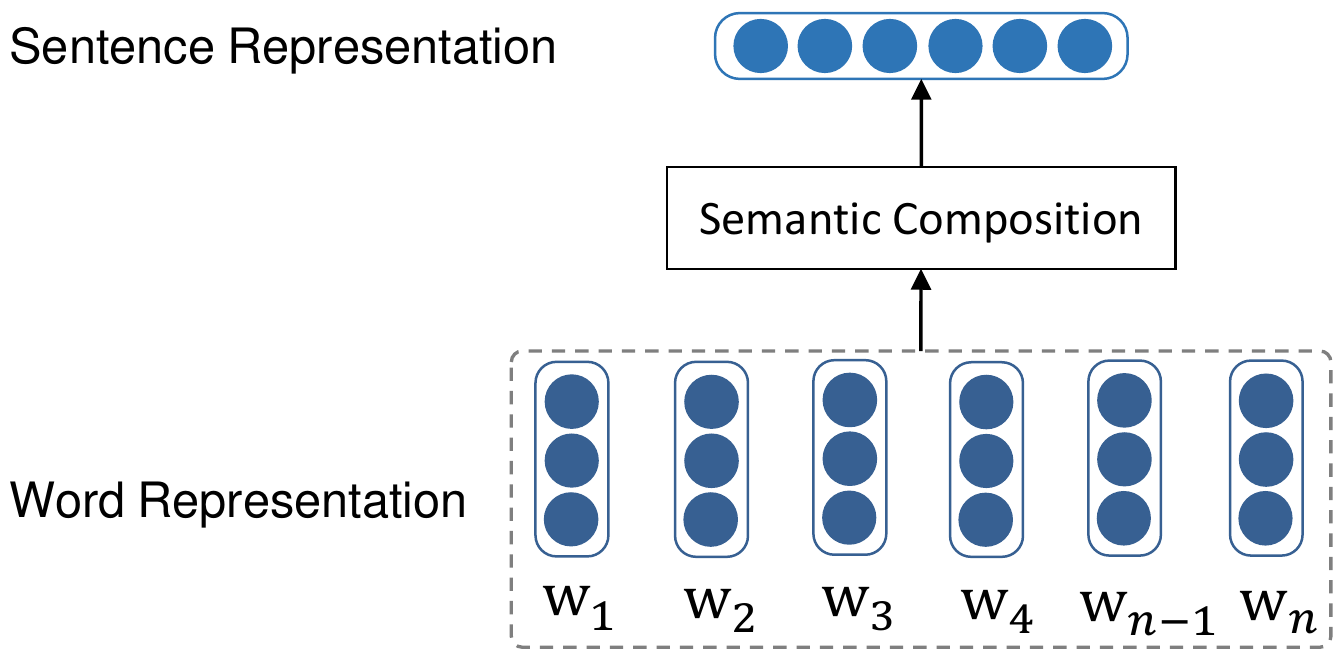}
	\caption{An illustration of semantic compositionality.}
	\label{fig:composition}
\end{figure}
For sentence $s$, we use semantic compositional method to calculate sentence representation from the representations of the word it contains.
This is on the basis of the principal of compositionality \cite{Frege1892}, which states that the meaning of a longer expression (e.g. a sentence) comes from the meanings of its constituents.
Since we do not have plenty of task-specific training data, unsupervised compositional approach is preferred to make the system scalable. As a case study, we also use average as the compositional function. We leave more sophisticated unsupervised approaches like Deep Boltzmann Machine \cite{Hinton2006}, Denoising Autoencoder \cite{Glorot2011}, Unfolding Recursive Autoencoder \cite{Socher2011a}, LSTM Encoder \cite{Li2015} as future work.

\paragraph{Extension 1}
It is worth noting that the purpose of sentence selecting is to obtain some ``fuel'' which can be used in the Sentence Organizing part to form an article. 
Based on this consideration, we believe that tagging each sentence with a discourse/semantic tag will help to organize the sentences with more evidences.
Therefore, we use an automatically discourse element labeling algorithm \cite{Song2015} to decide what role does a sentence acts as, such as ``Introduction'', ``Prompt'' and ``Conclusion''.
For example, ``Introduction'' sentence introduces the background and/or grabs readers' attention and ``Conclusion'' sentence concludes the whole essay or one of the main ideas.

\paragraph{Extension 2}
Given a list of words for one argument, sentence selecting part of outputs a collection of sentences where the input words come from the ``Topic Understanding'' part.
These selected sentences are semantically related to the argument, and might contain some supporting words which do not covered in the input word set. 
Based on this consideration, we add a ``feedback'' mechanism to extract some new words from the sentences and add them to the input word set as an expansion. In this way, the framework could work in a bootstrapping fashion.
We regard extracting new word $w$ from a collection of sentences $S$ as a ranking problem. What we need to do is designing a scoring function $f(w, W, S)$, where $w$ is a candidate word in the sentence collection $S$, $W$ is the input word set for one argument. After getting the score of each candidate $w \in S$, we rank them and select the top ranked $k$ words which are not contained in $W$. 
We explore two methods for extracting new words from the outputted sentences: a counting method and an embedding method. 
In counting based method, $f(w, W, S)$ is the number of $w$ occurs in $S$. 
In embedding based method, $f(w, W, S)$ is the similarity between  $vec_w$ and $vec_W$, where $vec_W$ is the averaging of vectors of words in $W$.



\subsection{Sentence Organizing}
In this part, we describe the sentence organizing part which organizes a set of chaotic sentences into an organized article. 
This can be considered as a structure prediction problem, and the objective is to predict a desirable structure of a collection of sentences.

\begin{figure}[h]
	\centering
	\includegraphics[width=.47\textwidth]{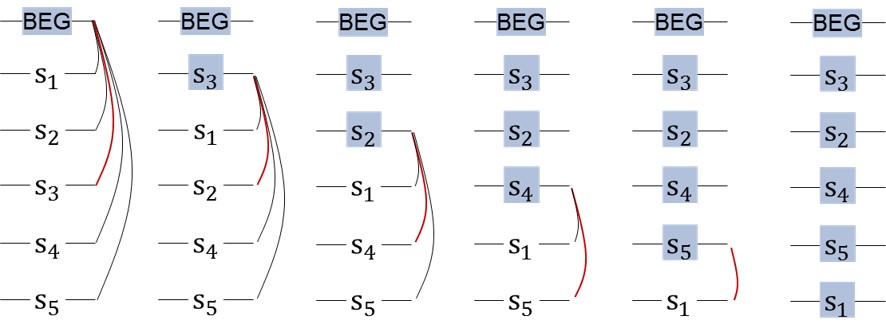}
	\caption{An illustration of the greedy method for sentence organizing. The red line means the sentence pair with highest semantic relatedness.}
	\label{fig:greedy-1}
\end{figure}

To this end, a natural choice is to greedily get the order of a list of sentence from left to right, one sentence at a time. That is to say, when we looking at a sentence, we only select which sentence is the most relevant one to be after the current sentence. This process could be done in a recursive way, so that a order could be generated greedily. An illustration of this idea is given in Figure \ref{fig:greedy-1}.
One important component in this setting is a scoring function $f(s_1, s_2)$ to weight the semantic similarity between two sentences $s_1$ and $s_2$. 

\begin{figure}[h]
	\centering
	\includegraphics[width=.4\textwidth]{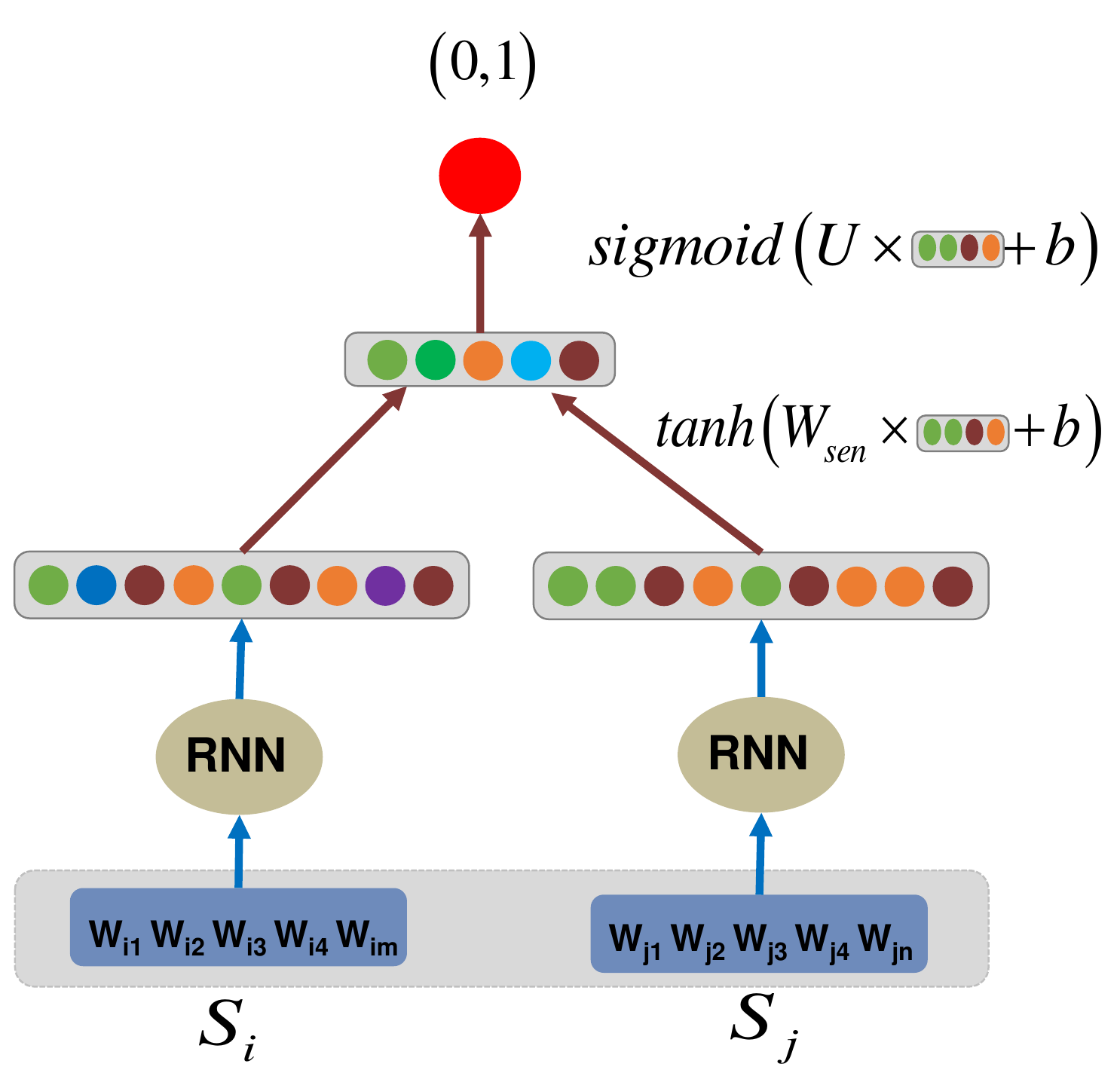}
	\caption{An illustration of the recursive neural network method for sentence scoring, where $s$ means a sentence and $w$ is a word.}
	\label{fig:rnn}
\end{figure}

We explore four methods as relatedness scoring function to calculate the coherence of two sentence. 	
\begin{itemize}
	\item Bag-of-Word (Boolean). We represent each sentence as bag-of-words, whose values represent whether a word occurs in the sentence. We use cosine similarity between $s_1$ and $s_2$ as their similarity.
	\item Bag-of-Word (Frequency). Similar with Bag-of-Word (Boolean), each sentence is represented in a bag-of-word fashion. The difference is that in this setting the values of each dimension is the frequency of each word occurs in a sentence.
	\item Embed-Average. We represent each sentence as a continuous vector and calculate the cosine similarity between two sentence vectors as $f(s_1, s_2)$. In this setting, the vector of each sentence is obtained by averaging the vectors of words a sentence contains. 
	\item Recursive Neural Network. We represent each sentence as a continuous vector, which is computed with recursive neural network \cite{Socher2011b}\footnote{One could also use recurrent neural network or convolutional neural network as alternatives.}. After that, we use a feed-forward neural network to score the relatedness between sentences $s_1$ and $s_2$. An illustration of this method is given in Figure \ref{fig:rnn}.
\end{itemize}

Among these four methods, the first three methods are similarity driven as they regard the cosine similarity between sentences as the scoring function. The last method is relatedness driven as there is an additional feed-forward neural network to encode the relatedness between sentences.
The first three models do not contain external parameters, while the fourth model needs to be trained from data.
In order to learn the parameters in recursive neural network and the the parameters in the feed-forward scoring function, we follow \cite{Li2014a} and use a cross entropy loss function. 
The basic idea is that during training the category of a correct pair of sentence $f(s_1, s_2)$ should be different from the category of a corrupted pair of sentence $f(s_1, s_2^*)$. 
We use cross entropy as the loss function \cite{Li2014a}, where $P_c(s_1, s_2)$ is the probability of predicting $(s_1, s_2)$ as class $c$ given by the $sigmoid$ layer, $P^g_c(s_1, s_2)$ indicates whether class $c$ is the correct category, whose value is 1 or 0. 
The correct pair of sentence is a real case in the corpus, namely there is a sentence $s_2$ occurring after sentence $s_1$.
The corrupted pair of sentence is artificially generally by replacing $s_2$ with a randomly selected sentence $s_2^*$. 

\begin{equation}
loss = -\sum_{s_i, s_j \in S}^{}\sum_{c=0,1}P_{c}^{g}(s_i, s_j) \cdot log(P_{c}(s_i, s_j))
\end{equation}

The greedy method mentioned above organized sentences in a local way.
That is to say, the method processes a sentence by only seeing its previous sentence, without capturing global evidence or optimizing a global organizing result.
To solve this problem, we use a global approach, which is also a natural choice inspired by the representative studies in sequence labeling like part-of speech tagging and named entity recognition \cite{Manning1999,Jurafsky2000}. 
The basic idea is that, when we deal with a potential sentence at index $i$, we consider the preceding histories and calculate the best score of a result from the beginning of a sentence to the current index.
We use dynamic programming to recursively calculate the scores and decode with standard Viterbi algorithm.  



\section{Experiment}
We compare between different methods for each component empirically. 
For ``Topic Understanding'' and ``Sentence Selecting'' parts, we only evaluate methods qualitatively as we do not have the ground truth. For ``Sentence Organizing'' part, we also evaluate different methods quantitatively as the original orders of sentences in a document could be viewed as a ground truth.
We conduct experiments on a Chinese dataset we crawl from the web, which contains 6,683 documents with 193,210 sentences.

\begin{CJK*}{UTF8}{gbsn}

\begin{table*}[t]\footnotesize
	\centering
	\begin{tabular}{c|p{14cm}}
		\hline
		Method & Top ranked results\\
		\hline
		Thes & 青年期 (adolescence)，青年 (young man)，妙龄 (sweet seventeen)，年青 (youthfulness)，年轻轻 (young)，青春年华 (the green years)，年轻 (young)，年幼 (childish)，韶华 (prime of youth)\\
		\hline
		TM& 激情 (passion)，活力 (vitality)，挥霍 (squander) ，时光 (days)，年少 (youth of age），无悔 (no regret)，充满 (suffused)，绚丽 (gorgeous)，短暂 (transient)，燃烧 (burn)\\
		\hline
		WE & 活力 (vitality)，激情 (passion)，豆蔻年华 (marriageable age)，花季 (flowering time)，无悔 (no regret)，朝气 (vigour)，年少 (youth of age)，轻狂 (frivolous)，青涩 (tender)，活力四射 (energetic)\\
		\hline
	\end{tabular}
	\caption{Results of different algorithms for topic expansion on youth (青春\ in Chinese).}\label{table:topic-understanding-1}
\end{table*}

\begin{table*}\footnotesize
	\centering
	\begin{tabular}{l|p{14cm}}
		\hline
		Method & Results\\
		\hline
		\multirow{10}{*}{Counting} 
		& In Chinese: 青春时节，深沉不属于我们，我们的青春，充满着对未来的希望与憧憬，张狂，是青春的代名词，我们，应当珍惜着美好的时节，去叩响生命的钟声。\\
		&Translation in English: When we are young, we should be full of hope to the future without many burdens. We can be wild. We should treasure such a beautiful age to live a happy life.
		\\
		\cline{2-2}
		& In Chinese: 青春就应该扬帆起航，不管遇到再大的风险，都不会使我们半途而废，因为我们始终坚信未来是美好的，未来一直在伸出双手迎接我们的到来，他在等着我们，他在等待着我们的成功。\\
		& Translation in English: People in youth could be full of coverage and keep moving no matter how many risks they come up with, as we always hold the faith that the bright future is reaching out her hands to embracing our arrival and is waiting for our success.\\
		\cline{2-2}
		& In Chinese: 青春是充满激情的生活，这是红花一样的年轮和绿叶一样的生命谱写的一首诗，一曲歌，一个梦，青春的诗篇豪迈奔放，青春的歌委婉动听，青春的梦想神气美好！\\
		& Translation in English:  Youth is a life full of passion, is a poetry composed with florid life, is a song and a dream. The poetry of youth is untrammeled, the song of youth is melodious, and the dream of youth is spirited and beautiful.\\
		\hline
		\multirow{6}{*}{Embedding} 
		& In Chinese: 在青春年华的岁月里将是充满希望、充满激情、充满奋斗的岁月。\\
		& Translation in English: The years of youth are full of hope, passion and endeavor. \\
		\cline{2-2}
		& In Chiense: 青春有志是书写我们青春无悔篇章的开端。\\
		& Translation in English: Great ambition in youth is the beginning to write a new regretless chapter. \\
		\cline{2-2}
		& In Chinese: 青春是不羁的资本！\\
		& Translation in English: Youth is the capital to be uninhibited! \\
		\hline
	\end{tabular}
	\caption{Results of different algorithms for sentence selecting on youth (青春\ in Chinese).}
	\label{table:sentence-selecting}
\end{table*}

\subsection{Evaluation on Topic Understanding}
We evaluate the effects of different algorithms for topic understanding. 
Since there are two steps in this part, we evaluate them separately. 
The results on youth (青春) are given in Table \ref{table:topic-understanding-1}, where Thes is thesaurus-based method, TM means topic model approach, WE is word embedding based method.
We uses Hownet\footnote{http://www.keenage.com/} as the external resource in Thes. 
The word embeddings used in WE are learned with Skipgram method \cite{Mikolov2013a}.
In thesaurus based methods, we filter out the words whose length are 1 because most of them do not have concrete meaning.
Despite using this filtering rule, we find that the results of Thes are still worse than others. The supporting words are formal, not commonly used in user generated articles.
Moreover, the meanings of supporting words are topic focused and do not go beyond the literal meaning of the input word. This is partly caused by the coverage of the thesaurus. 
We observe that the results of TM and WE are comparable and better than Thes in this example. For an noun ``youth'' (青春 \ in Chinese), TM and WE could find semantically related words which are more diverse and not restricted to the literal similarity. 
We take word embedding method as an example, and compare between K-Means and AP clustering algorithms to test their performances on ``Topic Clustering''. According to our observations, K-Means performs better than AP clustering as the results in AP contain many clusters containing less than 3 words.

\begin{table*}\footnotesize
	\centering
	\begin{tabular}{l|p{14cm}}
		\hline
       SId & Sentence Content\\
		\hline
		\multirow{2}{*}{1}
		& In Chinese: 未来是一张白纸，任时间任意涂抹，我宁愿步行，也不会去坐火车、汽车。 \\
		& Translation in English: Our future is blank, could be freely painted. I prefer walking to taking a bus or train.\\
		\hline
		\multirow{2}{*}{2}
		& In Chinese: 我要任意遨游，不要有轨道，不要有道路。 \\
		& Translation in English: I would like to find my future in my way, without some predefined restrictions.\\
		\hline
		\multirow{2}{*}{3}
		& In Chinese: 要去追寻自己的天地而不是去跟随他人的脚步。\\
		& Translation in English: I want to find my own way rather than following others' paths.\\
		\hline
		\multirow{2}{*}{4}
		& In Chinese: 因为我明白，成功永远不可能被复制。\\
		& Translation in English: Because I know that one's success could not be duplicated.\\
		\hline
		\multirow{2}{*}{5}
		& In Chinese: 人生亦不能。\\
		& Translation in English: It is also the truth for one's life. \\
		\hline
	\end{tabular}
	\caption{A randomly selected document used for observing the performances of different algorithms for sentence organizing on youth (青春\ in Chinese). SId means the original index of each sentence in the document. }
	\label{table:coherence}
\end{table*}

\subsection{Evaluation on Sentence Selecting}
We evaluate the performances of counting and embedding methods for sentence selecting. 
We also take youth (青春) as a case study.
The top selected results (in Chinese) are given in Table \ref{table:sentence-selecting}. 
We can find that the obtained sentences in counting based method are typically longer as it favors the sentences containing more key words. We believe that these results are more suitable to act as Prompt sentences because they include more specific evidences.
On the contrary, the results of embedding based method are typically shorter and more cohesive, which is partly caused by the way we used for composing sentence vector. 
Such results might be regarded as Theme or Conclusion sentences which are more abstractive.
For both methods, it is somewhat disappointing that they favour to selecting the sentences containing topic words such as ``youth'', which is less diverse than we have expected.
\begin{table}[h]\footnotesize
	\centering
	\begin{tabular}{l|c}
		\hline
		Method & Sentence Order\\
		\hline
		Ground Truth & 1 2 3 4 5\\
		bow binary & 1 2 4 3 5\\
		bow freq & 1 2 3 5 4\\
		bow embed & 1 2 3 5 4 \\
		Rec NN & 1 2 3 4 5\\
		\hline
	\end{tabular}
	\caption{Results of different algorithms for sentence organizing on youth (青春\ in Chinese). }
	\label{table:coherence-results}
\end{table}


%
%
%
%
%
%
%
%

\subsection{Evaluation on Sentence Organizing}
In this part we use two experimental settings to compare between different four coherence functions.
In the first setting, we take greedy framework as a case study and qualitatively evaluate them in the real system by showing the sentence orders generated from different coherence functions, including BOW (Boolean), BOW (Frequency), Embedding (Avg) and Recursive NN. 
In the second setting, we quantitatively evaluate them on a hold-out dataset consisting of several documents. The input for each coherence model is the same, namely the sentences of a document and the first sentence. The output is the orders generated from each coherence model. As we have the original sentence order, we can regard it as the ground truth, and evaluate the quantitative performance in terms of accuracy on bigrams of sentences, which is similar with Bleu \cite{papineni2002bleu} in machine translation and Rough \cite{lin2004rouge} in summarization.

An illustration of the different sentence organizing results are given in Table \ref{table:coherence} and Table \ref{table:coherence-results}. We choose a relatively short document consisting of five sentences to show the performance. We can find that recursive neural network performs gets the correct order while all others have one or more mistakes.
Different from the previous two components, we can quantitatively evaluate the performance of this part as the original sentence order of a document could be regarded as the ground truth of the sentences it contains.

\begin{table}[h]\footnotesize
	\centering
	\begin{tabular}{l|c|c}
		\hline
		Method & Greedy & DP\\
		\hline
		BOW (Boolean) & 0.213 & 0.217\\
		BOW (Frequency) & 0.194 & 0.209\\
		Embedding (Avg) & 0.233 & 0.238\\
		Recursive NN & 0.241 & 0.257 \\
		\hline
	\end{tabular}
	\caption{Results of different algorithms for sentence organizing on youth (青春\ in Chinese).  DP means dynamic programming.}
	\label{table:sentence-organizing}
\end{table}
\end{CJK*}
Experimental results are given in Table \ref{table:sentence-organizing}.
We can find that the performances of the four methods in greedy and DP setting are consistent. 
BOW (Boolean) outperforms BOW (Frequency) consistently. This indicates that whether a word occurs in two sentences could indicate the relatedness of two sentences and it does not need to consider how many times a word occurs in each sentence. 
Average based embedding method performs better than bag-of-word methods by considering the continuous word and sentence representation in some latent semantic space. 
We can find that recursive neural network method performs best in each setting, outperforming three previous similarity based methods. This shows the effectiveness of a powerful semantic composition model as well as the necessary to model the relatedness between two sentences rather than a cosine based similarity measurement.

\section{Related Work}
We briefly talk about some related works in literature about natural language generation and essay generation in this section. 

Natural language generation (NLG) is a fundamental and challenge task in natural language processing and computational linguistics \cite{Manning1999,Jurafsky2000,Reiter2000}.
The task of essay generation could be viewed as a special kind of natural language generation. 
Existing NLG approaches could be divided into three categories: template based methods, grammar based methods and statistical based methods.
Template based methods typically use manually designed templates with some slots and replace words to generate new article.
Grammar based methods go one step further by manually designing some structured templates and compose an article with computer program.
Statistical based methods focus on learn the sophisticated patterns from the web and generate article in an automatically way.
In this work, we follow \cite{Reiter1997} and explore planning based approach. 
There also exists some related studies in text generation.
For example, Belz \shortcite{Belz2007} generate weather report texts with probabilistic generation method.
Jiang and Zhou \shortcite{Jiang2008} generate Chinese couplets with statistical machine translation approach. 
Angeli et al. \shortcite{Angeli2010} develop a domain-independent method with a sequence of local decisions, and evaluate the method on Robocup sportscasting and technical weather forecasts.
Zhang and Lapata \shortcite{Zhang2014b} generate Chinese poetry with recurrent neural network. 
Shang et al. \shortcite{Shang2015} generate short-text conversation with neural responding machine.
Li et al. \shortcite{Li2015} generate a paragraph/document with attention based neural network. 
Rush et al. \shortcite{Rush2015} and Hu et al. \shortcite{Hu2015} use attention based recurrent neural network to generate abstractive summarization. 
Image captioning \cite{xu2015show} can also be viewed as a kind of text generation which takes a picture as the input. 

\section{Conclusion and Future Directions}
In summary, we develop a planning based framework to generate an article by taking a topic word as input. The framework consists of three components: a topic understanding component, a sentence selecting component and a sentence organizing component. 
We also add a feedback mechanism to enhance the results of topic understanding. 
For each component, we explore several methods and conduct a case study on a Chinese corpus.
We show that for topic understanding, topic model and word embedding based methods perform better than thesaurus based methods.
Recursive neural network based model performs better than bag-of-word and embedding average based similarity driven methods for sentence organizing.

There remains plenty of challenges in this line of research.
One direction is how to quantitatively evaluate the effectiveness of each internal component as well as the final generated article. In this work, we only quantitatively evaluate the sentence organizing part as the original sentence order could be used as gold standard. However, for other parts, it is impractical to build gold standard for each input topic word. It is desirable to find some automatic evaluation methods or to test the algorithms on some applications with automatically labeled gold standards.
For the sentence selecting part, we find that the methods we tried prefers to select the sentences which contains the exact supporting words. The task requires us to choose more diverse sentences with different semantic roles to form a diversified document.
From another perspective, the input of this work is a given topic word. However, in some situation there is only an idea or a description about what we want to write. We need to understand the idea/description and get the topic word. We leave this as another potential future work.
Furthermore, the methods used in this work can be regarded as an extractive approach. 
We separate the whole task into several subtasks and develop algorithms to address each part. 
This might suffers from the problem of error propagation, and this could be to reduced to some extend if we build an end-to-end method like the emerging neural network approach. This is also a very interesting future work. 

\bibliography{bibtex}
\bibliographystyle{naaclhlt2016}

\end{document}